# Code-Switching and Syntax: A Large-Scale Experiment

Igor Sterner* and Simone Teufel
Department of Computer Science and Technology
University of Cambridge
United Kingdom
{is473,sht25}@cam.ac.uk

## Abstract

The theoretical code-switching (CS) literature provides numerous pointwise investigations that aim to explain patterns in CS, i.e. why bilinguals switch language in certain positions in a sentence more often than in others. A resulting consensus is that CS can be explained by the syntax of the contributing languages. There is however no large-scale, multi-language, cross-phenomena experiment that tests this claim. When designing such an experiment, we need to make sure that the system that is predicting where bilinguals tend to switch has access only to syntactic information. We provide such an experiment here. Results show that syntax alone is sufficient for an automatic system to distinguish between sentences in minimal pairs of CS, to the same degree as bilingual humans. Furthermore, the learnt syntactic patterns generalise well to unseen language pairs.

## 1 Introduction

Language contact occurs when language communities intersect. One product of nearly all language contact situations is the phenomenon of code-switching (Gardner-Chloros, 2020). Code-switching (henceforth CS) occurs when multilingual speakers use more than one language in an utterance.

It is generally agreed that bilinguals have shared knowledge of what constitutes acceptable CS (Joshi, 1982). However, every CS model ever proposed has been refuted in subsequent investigations, which presented data that the model is unable to explain (Clyne, 1987; Mahootian, 1993; Auer and Muhamedova, 2005; Poplack and Cacoullos, 2016; Nguyen, 2024). Computational experiments have aimed at the comparison of existing theories, but still no one single theory has emerged (Pratapa and Choudhury, 2021).

*The author is now at the University of Edinburgh.

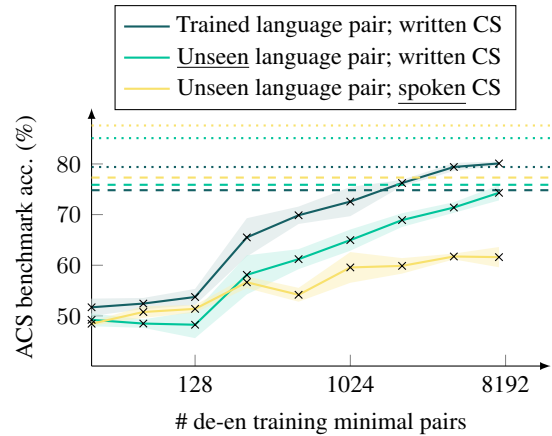

Figure 1: Syntax-based model performance on the ACS benchmark for increasing number of training samples. Dashed lines represent `Llama-3.1 405B` performance, dotted lines represent human performance.

One general consensus of these decades of theoretical study is that the syntactic structure of the CS utterances are a good starting point for any future theory (MacSwan, 2000; Auer and Muhamedova, 2005; Nguyen, 2024). This postulate can be traced back to the equivalence constraint (Poplack, 1980), a claim that CS can only occur at points when the surface structure of the contributing languages align. This precise theoretical claim has since frequently been challenged (see e.g. Mahootian, 1993; Eppler, 2005), but no large–scale empirical evidence has been brought forward.

In the present work, we do not aim to compare existing theories of CS or propose any new theory of CS. Instead, our goal is to empirically answer the question: *Is syntactic structure alone sufficient to distinguish naturally-observed from synthetically manipulated CS?* Any future theory of CS will only stand the test of time if it can be successfully applied across a wide range of data sources and language pairs. We will therefore also aim to empirically answer the question: *To what extent do syntactic CS patterns in one language pair gen-*

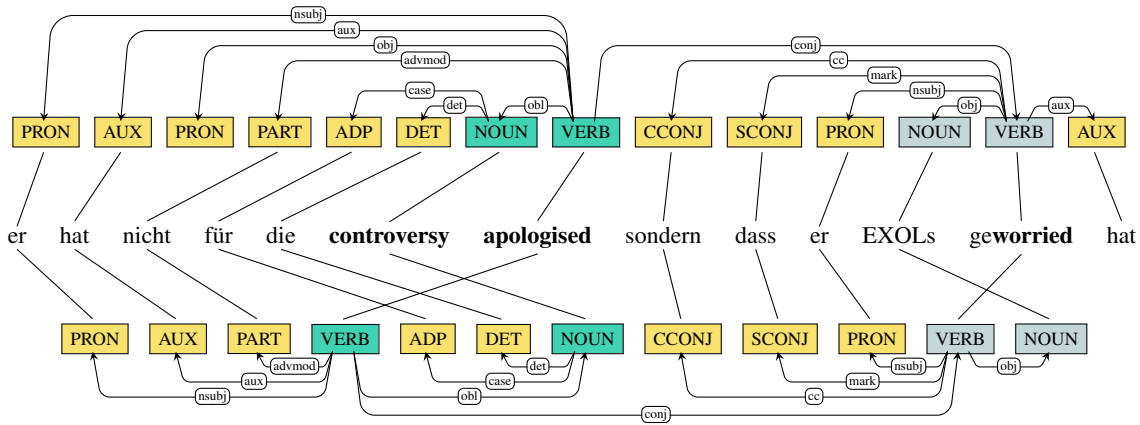

Figure 2: Example input to the GNN. Original CS sentence and alignments provided only for reference. Yellow is German, green is English and grey is named entities, words of mixed morphology or punctuation.

*eralise to others?* A consideration in this second question is the domain of CS under investigation; research has shown that spoken CS has both similarities and differences to written CS (Gardner-Chloros and Weston, 2015). Gardner-Chloros and Edwards (2004) state that written CS has stronger grammatical constraints than spoken CS, whereas Chi and Bell (2024) provide evidence that may suggest otherwise.

In previous work (Sterner and Teufel, 2025), we designed an evaluation methodology and benchmark that makes it possible to answer these questions. Our benchmark contains up to 1,000 minimal pairs of CS sentences each for 11 language pairs. As part of the validation of that benchmark, a sample of minimal pairs was judged by at least one bilingual for all language pairs.

The construction of the minimal pairs involves the translation of a CS sentence into two monolingual versions, as shown in Figure 3. This makes it possible to run a universal dependency parser (Nivre et al., 2017). Additionally, CS information is retained by mapping word-level language identification onto the words in the monolingual translations. A minimal pair is then algorithmically created by changing the language of one word at the switch point. In the previous work, we tested off-the-shelf language models on the task of predicting how relatively acceptable a sentence is compared to its counterpart in a minimal pair.

In this paper, we present a trained computational model for the same task. It is neuro-symbolic. As input, instead of the surface forms of the sentences, it receives the syntactic analyses of the two monolingual versions. Good performance of such a model would be empirical evidence that syntax alone can predict where bilinguals are likely to switch language in a CS sentence. Furthermore, we will train the model only on German–English written CS, and test it on data from the same language pair, other language pairs of written CS and language pairs of spoken CS. Figure 1 gives a summary of results for 3 language pairs (written German–English CS, written Spanish–English CS and spoken Turkish–German CS); they are representative for many other language pairs for which we do not have space here (see Appendix Figure 4). We can see that training based on syntax leads to good performance on the benchmark. In order to approach the human ceiling, training data on the order of 6,000-8,000 pairs is necessary. Further, the learnt patterns generalise well to written CS data from an unseen language pair. When transcriptions of spoken CS are used instead of written CS, generalisation still holds but is weaker. The results provide concrete empirical evidence – on a scale much larger than previous experiments – that general syntactic patterns in CS do exist.[1]

## 2 Computational Model of CS

We design a model that is trained to take as input syntactic information from the contributing languages of a CS sentence and output a prediction of the more acceptable CS sentence in a minimal pair.

### 2.1 Design

We train a graph neural network (GNN). The input we provide to the model is only syntactic information in the form of universal dependencies (UD;

---

[1]Code is available at https://github.com/igorsterner/csntax-gnn.

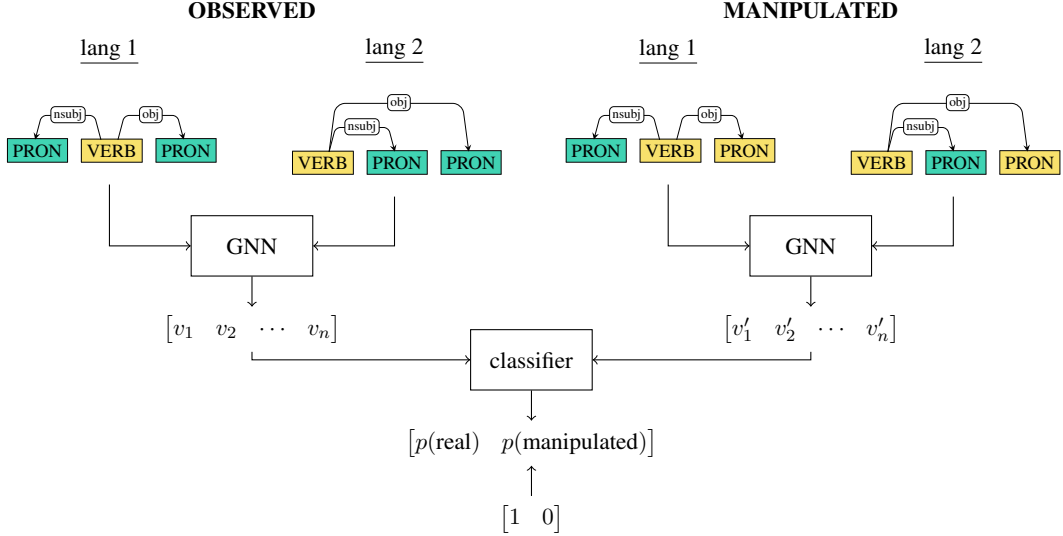

Figure 3: The training regime of `CSntax-GNN`.

Nivre et al., 2017); UD was chosen because it provides a consistent syntactic representation across all languages. In UD dependency parse structures, words constitute nodes and edges between nodes represent directed dependencies. Nodes in our representation have two features: the POS of the word, and the language of the aligned word in the CS sentence.

Each sentence in a minimal pair is represented by two graphs, one for each monolingual translation. These are passed as unconnected graphs to the GNN, which generates a single embedding from them. For each alternative sentence in a minimal pair, one such embedding is created. An MLP classifier then learns to compare these two embeddings. All parameters of the GNN and the classifier (a total 2.7K) are updated in an end-to-end manner with cross-entropy loss. The training setup of our model is shown in Figure 3. We name the resulting system `CSntax-GNN`.

### 2.2 Experimental Setup

**Architecture** We instantiate `CSntax-GNN` as a graph isomorphism network (GIN, Xu et al., 2019). In order to model dependency edges with features, we use the adaptations proposed by Hu et al. (GINE, 2020). The final operator used for each layer is as follows:

$$\mathbf{x}'_i = h_{\boldsymbol{\theta}} \left( (1 + \epsilon) \cdot \mathbf{x}_i + \sum_{j \in \mathcal{N}(i)} \mathrm{ReLU} \left( \mathbf{x}_j + \mathbf{e}_{j,i} \right) \right)$$

where $\mathcal{N}$ contains the nodes (the words), $x$ represents node features, $e_{j,i}$ is the edge feature between node $j$ and node $i$ and $x'$ is the updated node features after the layer. $\epsilon$ is a learnt constant and $h$ corresponds to the learnable network with weights $\theta$. In the model, each of these networks is realised as an MLP.

We append a binary feature to the nodes representing whether they come from the graph of Lang1 or Lang2 and also add self-connections (Kipf and Welling, 2017). We use three layers and train with a constant learning rate = 0.001 in batches of 128 graphs.[2]

**Data** We use the train, validation and test sets provided by the ACS benchmark and corpus (Sterner and Teufel, 2025). There are 7,501 training pairs (from tweets in 04/2019–12/2021), 1,932 validation pairs (from tweets in 01/2022–06/2022), and 1,000 testing pairs (from tweets in 07/2022–02/2023). All pairs contain two sentences, which were automatically segmented from tweets from existing corpora (DeLucia et al., 2022; Sterner and Teufel, 2023; Frohmann et al., 2024). We use the provided tokenization (which is from Qi et al.'s tools (2020)) and token-based language identification (which is from Sterner's tool (2024)).

For fairness, we need to repeat the subsequent analysis for every sentence independently. Sentences are translated monolingually into both contributing languages using the `madlad-3B` NMT model (Kudugunta et al., 2023); word-to-word alignments are accomplished by the `awesome-align` model (Dou and Neubig, 2021); and the monolingual sentences are transformed into universal dependencies using `stanza` (Qi et al., 2020).

---
[2]Training `CSntax-GNN` is performed on a single node with three CPU cores and 32 GiB of RAM. Training takes approximately 2 minutes.

|  | de-en | da-en | es-en | fr-en | it-en | id-en | nl-en | sv-en | tr-en | tr-de | zh-en |
|---|---|---|---|---|---|---|---|---|---|---|---|
| Human | 79.4$_{3.1}$ | 83.1 | 85.1 | 75.1 | 86.6 | 86.1 | 81.1 | 82.1 | 85.6 | 87.6 | 79.6 |
| `Llama-3.2 1B` | 62.9 | 66.7 | 64.1 | 63.0 | 65.1 | 63.7 | 64.1 | 64.6 | 60.9 | 61.6 | **66.2** |
| `Llama-3.2 3B` | 66.6 | **69.4** | 66.9 | 65.6 | 69.4 | 68.1 | 69.4 | 67.1 | 63.8 | 61.6 | 65.7 |
| `Llama-3.1 8B` | 68.3 | **71.8** | 69.1 | 69.0 | 70.3 | 72.8 | 69.5 | 72.0 | 66.4 | 68.5 | 71.1 |
| `Llama-3.1 70B` | 73.2 | **74.2** | 73.0 | 72.3 | 73.9 | 77.8 | 73.9 | 75.0 | 70.9 | 75.0 | 70.6 |
| `Llama-3.1 405B` | 74.8 | **74.5** | 75.9 | 75.4 | 75.5 | 79.7 | 75.6 | 76.2 | 74.3 | 77.3 | 67.7 |
| `XLM-R base` | 63.7 | **69.7** | 60.8 | 60.8 | 61.5 | 71.4 | 66.1 | 70.1 | 60.4 | 63.9 | 61.7 |
| +FT | 62.3$_{0.2}$ | 56.1$_{0.9}$ | 59.5$_{0.6}$ | 53.8$_{0.3}$ | 60.6$_{1.0}$ | 62.1$_{0.5}$ | 58.0$_{0.5}$ | 61.2$_{0.5}$ | 55.0$_{0.6}$ | 56.9$_{1.4}$ | 60.9$_{0.2}$ |
| `XLM-R large` | 68.5 | **73.6** | 61.7 | 64.5 | 67.0 | 75.6 | 70.4 | 73.6 | 64.9 | 66.2 | 59.7 |
| +FT | 64.5$_{1.0}$ | 59.0$_{0.9}$ | 58.2$_{1.3}$ | 58.5$_{0.3}$ | 61.4$_{1.7}$ | 63.9$_{1.2}$ | 60.4$_{0.6}$ | 62.5$_{0.7}$ | 58.8$_{0.9}$ | 54.5$_{0.8}$ | 59.0$_{0.8}$ |
| `CSntax-GNN` | **79.4**$_{0.5}$ | **74.3**$_{0.8}$ | **72.1**$_{1.7}$ | **74.7**$_{0.7}$ | **76.3**$_{0.8}$ | **76.0**$_{1.0}$ | **72.4**$_{1.0}$ | **73.7**$_{0.4}$ | **73.0**$_{1.8}$ | 62.8$_{2.1}$ | **66.7**$_{1.5}$ |

Table 1: Model performance on the ACS benchmark. Boldface indicates the best group of models for each language pair. All bold-faced values are statistically indistinguishable from each other. Subscript numbers give standard deviation for non-deterministic systems.

**Baselines** As our first automatic baseline, we use the `Llama-3` results listed in (Sterner and Teufel, 2025) for the ACS benchmark. They are calculated based on sentence probabilities.

The training data available to `CSntax-GNN` was generated using the same procedure as the test data. This is of concern to us, because we want to be able to attribute good performance to the model's use of its input syntactic information and not to specifics of the data generation. Hence our second baseline will be a language model finetuned with the same full set of training data available to `CSntax-GNN`. Our choice of `XLM-R` base and large models (Conneau et al., 2020) for these experiments is based on the fact that they have been widely shown to perform well in sequence classification tasks. If this finetuned model performs well, we expect it has found spurious features in the data. The setup for `XLM-R+FT` is identical to `CSntax-GNN`, except for the source of the embeddings (from mean pooling of node features for CSntax-GNN and from CLS representations for XLM-R+FT). The difference between the two therefore directly measures the contribution of syntactic information as compared to lexical semantic information.

We train the base model for 40 epochs and the large model for 20 epochs. There is a constant learning rate of 1e-4 and a batch size of 128 minimal pairs. Performance curves are provided in Appendix Figure 5.

| Ablated setting | Acc (%) |
|---|---|
| Baseline | 79.4$_{0.5}$ |
| (1) Random dep relation types | 79.1$_{0.3}$ |
| (2) Random POS tags | 77.7$_{0.8}$ |
| (1) + (2) | 73.5$_{0.1}$ |
| (3) Random language IDs | 56.8$_{0.7}$ |
| (1) + (2) + (3) | 51.1$_{0.5}$ |
| GAT architecture | 75.7$_{0.4}$ |

Table 2: `CSntax-GNN` ablation study

**Metrics** Our metric is accuracy (A). We test significance using the two-tailed paired Monte-Carlo permutation test with $R = 10,000$ and $\alpha = 0.05$. For significance testing of non-deterministic systems, we use the median run of three. For comparisons between human and system judgments, we use an unpaired version of the same test.

### 2.3 Results

Results are given in Table 1. If tested on the same language pair as it was trained on, German–English, `CSntax-GNN` reaches A=79.4%, which is indistinguishable from human performance (p=0.293). In this configuration, `CSntax-GNN` beats all baselines, including the largest pre-trained LLM (p=0.012), which was measured at A=74.8%. Furthermore, the finetuned variant of XLM-R performs badly (base A=62.3%, large A=64.5%). This means that there is no evidence that `CSntax-GNN` is able to exploit the data generation procedure in an unfair manner. Overall, these results provide concrete evidence that it is the syntactic representation available to `CSntax-GNN` that enables its good performance.

Table 2 gives the results of an ablation study on the model. We investigate which of the avail-able information – POS tags, dependency type labels, relational dependency structure, GNN architecture – is most important to its good performance. Randomising the POS features or dependency relation types results in a statistically insignificant change (p=0.322, p=0.320). When we randomize both, there is an absolute drop in performance of 6.6%, which is significant (p=0.031). This result confirms that it is the relational structures of the dependency parse that enable CSntax-GNN's good performance. Our choice of architecture (GINE) is the most general flavour of messaging-passing GNNs, because messages are directly learnable. We also ran another architecture, GAT (Velikovi et al., 2018), where the weighting between neighbouring messages is learnt. The GAT architecture results in a significant drop in performance of 4.4% (p=0.031).

Randomising the token-level language features should yield random performance, but results show that it is 6.8% above random. This is an indication of the best possible performance attainable by exploiting the data generation procedure. If we further randomise the language features such that the model can only exploit length and graph-based structural biases, results revert almost to random (A=51.1%).

For the 10 other language pairs, there is no training data available. Results for these show a gap between CSntax-GNN and human performance. But for the 8 language pairs where data was derived from social media, the model still performs numerically well (A=69.7-76.9%). This is empirical evidence that syntactic patterns of CS in German–English generalise to many other language pairs. For two of the other language pairs, Turkish–German and Chinese–English, the data originated from transcriptions of spoken recordings. The model performs worse for them, suggesting that syntactic patterns in spoken corpora differ from those in online social media.

We performed an error analysis on the automatic tools used. For this, the second author of this paper analysed 100 German–English minimal pairs and the output of the tools.[3] Results showed that 84 sentences were genuine CS sentences. Of these, 70% of all pairs were treated perfectly by all tools. In sum, the error analysis shows that there is noise in the training signal available to CSntax-GNN, but our main results show that it is still useful training material for the model.

Meanwhile, some error cases in the data do not propagate to the GNN. This is because the GNN does not take lexical information as input, only POS tags. This makes the task harder for the model, because the POS masks the lexical error. For example, in one minimal pair `a` `neuer` ('a new') becomes fully-German `einen neuer`. This results in ungrammatical adjective definiteness and therefore ungrammatical German morphology, something that an LLM can easily detect. But to CSntax-GNN, these two words are simply a determiner and an adjective. As a result, CSntax-GNN is not able to benefit from the obviously implausible second sentence in the minimal pair.

Both the best LLM and CSntax-GNN have achieved good performance on the benchmark. There were only 4.4% cases where neither system predicted the correct answer. In other words, they do not agree well with each other in their decisions ($\kappa$=0.28, N=1000, n=2, k=2). Evidently, both syntax and lexical semantics are important contributors to CS acceptability. This may motivate a hybrid system, for example augmenting node features with language model representations.

We investigated to what extent the certainty in the predictions of CSntax-GNN are correlated with human agreement on the task. In previous work, we found that probabilities from large LLMs are correlated with human agreement (Sterner and Teufel, 2025). However, for CSntax-GNN we found no such correlation even though we experimented with various approaches, including temperature-scaled variants. We can only conclude that CSntax-GNN may not make judgements of acceptability in a way that aligns with human CS language production.

## 3 Conclusion

The current experiment provides concrete empirical evidence – on a scale much larger than all previous experiments – that general syntactic patterns influence the way in which CS operates. We propose CSntax-GNN – a GNN trained on a syntactic representation of minimal pairs that is universal across languages. On the same language pair as it is trained, it achieves performance comparable to bilingual humans. We also posed the question to what extent syntactic patterns in one language pair generalise to others. For unseen written language pairs we observe a drop in performance (avg.

---

[3]A visualisation similar to Figure 2 was provided, except the actual words in the translation are provided, rather than universal dependency formalism.

-5.3%), but overall performance is still good. This shows that the syntactic patterns the model learnt on German–English do generalise to unseen language pairs.

## Limitations

The generalisation of CSntax-GNN to other language pairs that we reported here is biased towards its German–English training data. Our study was focused on German–English because of data and tool availability. This is a high resource resource language pair. For other language pairs, tools perform worse and we found the quality of the minimal pairs is lower. We hope that the methodology we present here can be applied by others to other language pairs.

Our operationalisation models CS as an interaction between two static grammars. This is different from a notion of a converged grammar (Clyne, 1987) that bilinguals may use. We have also not trained our model to learn any notion of the proficiency of the speaker in each of the contributing languages.

The ACS benchmark contains isolated sentences in minimal pairs. This may bias the comparisons towards syntactic factors, as compared to the wider context or speaker-specific contributions.

## Acknowledgements

The first author was supported by Cambridge University Press & Assessment and a Cambridge Trust studentship supported by Pembroke College, Cambridge.

## A Complementary Results

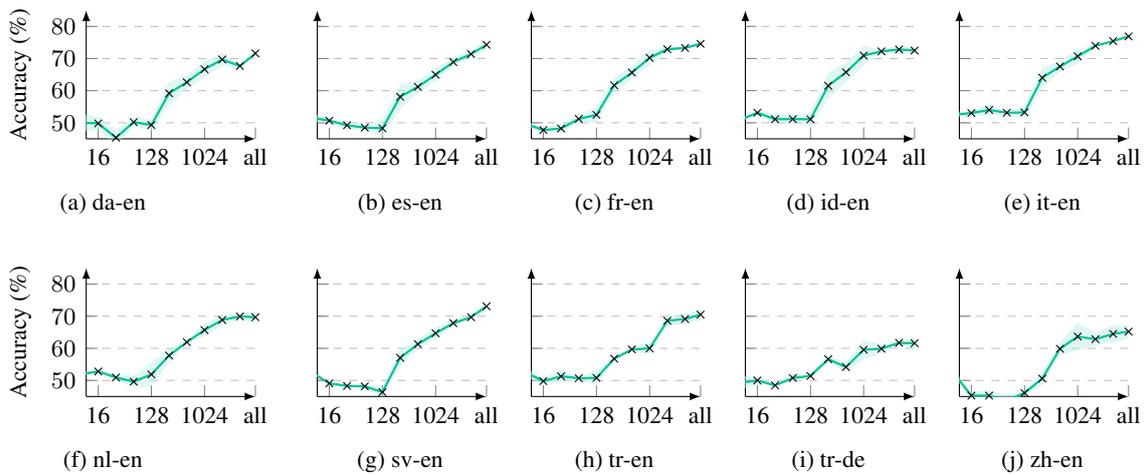

Figure 4: Number of German–English training minimal pairs against accuracy of `CSntax-GNN` on other language pairs.

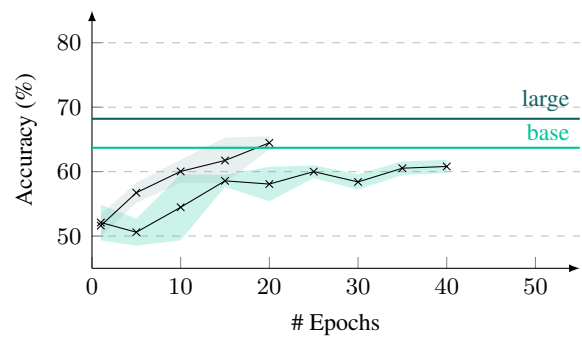

Figure 5: `XLM-R+FT` performance curves. Horizontal lines indicate training-free accuracies using pseudo-probabilities from the model directly.